\documentclass[journal]{IEEEtran}
\usepackage{cite}

\usepackage{hyperref}
\hypersetup{colorlinks,citecolor=blue,linkcolor=[rgb]{0,0,0.3},urlcolor=[rgb]{0,0,0.3},pdftitle={SI paper}}

\ifCLASSINFOpdf
  \usepackage[pdftex]{graphicx}
  \graphicspath{{Figures/},{Figures/Results/},{Figures/Systems/}}
   \DeclareGraphicsExtensions{.pdf,.jpeg,.jpg,.png,.eps}
\else
\fi
\usepackage{wrapfig}

\usepackage{amsmath}
\usepackage{amssymb}
\newcommand{\myMethod}{SIM}
\newcommand{\myMethodFull}{Structural Identifiability Mapping}

\usepackage{pifont}

\usepackage{diagbox}
\usepackage{tabularx,multirow} 

\usepackage[usenames,dvipsnames]{color}
\definecolor{RoyalBlue}{cmyk}{1, 0.50, 0, 0}
\newcommand{\new}[1]{\textcolor{black}{#1}}      
\newcommand{\kb}[1]{\textcolor{black}{#1}}      
\newcommand{\eco}[1]{\textcolor{black}{#1}}     
\newcommand{\pt}[1]{\textcolor{black}{#1}}      
\newcommand{\mjc}[1]{\textcolor{black}{#1}}     

\begin{document}

\title{Structure is information: structural identifiability mappings for machine learning with partially
observed dynamical systems}
%
%
%


\author{Janis~Norden,
        Elisa~Oostwal,
        Michael Chappell,
        Peter Ti\v{n}o
        and Kerstin Bunte
\thanks{J. Norden is with the Bernoulli Institute for Mathematics, Computer Science and Artificial Intelligence, University of Groningen, 
The Netherlands, e-mail: j.norden@rug.nl}
\thanks{\copyright 2026 IEEE.  Personal use of this material is permitted.  Permission from IEEE must be obtained for all other uses, in any current or future media, including reprinting/republishing this material for advertising or promotional purposes, creating new collective works, for resale or redistribution to servers or lists, or reuse of any copyrighted component of this work in other works.}
\thanks{Published version DOI: 10.1109/TCYB.2026.3694145}}

%
%

\markboth{IEEE TRANSACTIONS ON CYBERNETICS}%
{Norden \MakeLowercase{\textit{et al.}}: Time Series Classification in the Space of Structurally Identifiable Parameter Combinations}

%



\maketitle


\begin{abstract}
    The successful application of modern machine learning for time series classification is often hampered by limitations in quality and quantity of available training data.    
    To overcome these limitations, domain knowledge can be leveraged in the form of parameterised mechanistic dynamical models, whereby time series observations may be represented as instances of a predefined class of dynamical systems.
    \new{Provided the dynamical models are interpretable in terms of domain-specific variables \pt{and their dynamic interaction}, the learning process becomes interpretable as well and enables the modeller to handle sparsely and irregularly sampled data naturally.}
    However, the internal processes of a dynamical model are often only partially observed.
    This can lead to ambiguity regarding which particular model \kb{realization} best explains a given time series observation.
    This problem is well-known in the literature, and a dynamical model with this issue is referred to as structurally unidentifiable.
    Training a classifier that ignores knowledge about a structurally unidentifiable dynamical model can negatively influence classification performance.
    To address this issue, we employ structural identifiability analysis to explicitly relate parameter configurations that are associated with identical system outputs.
    Using the derived relations in classifier training, we demonstrate that this method significantly improves the classifier's ability to generalize to unseen data on a number of example models from the biomedical domain.
    This effect is especially pronounced when the number of training instances is limited.
    Our results demonstrate the importance of structural identifiability, a topic that has received relatively little attention from the machine learning community.
\end{abstract}


%
\IEEEpeerreviewmaketitle

\section{Introduction}

\IEEEPARstart{T}{he} problem of time series classification concerns the assignment of an observed time series to one of a predefined set of classes. 
As time series data naturally arise in a wide variety of scientific disciplines, including medicine, engineering and the social sciences, the associated theory of classification has been applied with great success to a multitude of problems such as health monitoring~\cite{mikalsen2016learning}, maintenance of civil infrastructure~\cite{carden2008ARMA} and emotion recognition from speech~\cite{trentin2015pattern}. 
While research has been directed at this problem \mjc{over many years}, reliable and time-efficient classification of time series 
remains a challenging problem to date \cite{pei2018multivariate, bianchi2021reservoir, Bradde2021multiclass}.

In many disciplines, the classification task is accompanied by domain-specific knowledge in the form of a mechanistic model which describes the data-generating mechanism.
This kind of model is typically derived from the application of a physical, chemical, biological or sociological law, and often takes the form of a parametrised dynamical system model.
If such a model is available, then its incorporation into the classification task is beneficial in two ways: 
Firstly, interpreting the data in the context of the mechanistic model makes it possible to deal with sparsely and irregularly sampled time series data in a natural way.
Secondly, any learned classification rule becomes interpretable as it directly relates to the given mechanistic model \cite{shen2017classification}.  
The second point is of particular importance for the modelling of high-risk applications common to biomedical and engineering domains, where the interpretability of \mjc{the application of machine learning} is of critical importance.
When safety, correctness \mjc{and trustworthiness} need to be guaranteed, model-based approaches are often the only viable option.

\new{Model- and data-driven hybrid techniques, often referred to as scientific or physics-informed machine learning, have emerged to leverage the scientific knowledge in order to analyse, simulate, and predict the behaviour of complex physical, biological, or engineering systems, with a recent review provided in \cite{Meng2025PhysicsMeetsML}. 
This typically includes the use of physical laws, constraints, or symmetries, which are embedded into machine learning models to improve data efficiency, interpretability, and generalization, especially in regimes with limited or noisy data \cite{zheng_sparse_2024,kamath_physics-informed_2024,ping_parameter_2025}. 
Beyond prioritizing accuracy, biological plausibility and realism are increasingly becoming central objectives in time series analysis \cite{yuan_dynamic_2025}.}
Assuming that a mechanistic model in the form of a parametrised dynamical system model is indeed available, \mjc{then} it is not automatically guaranteed that model-based classification works well.
The degree to which variables in the dynamical model can be observed is often restricted by practical and ethical considerations. 
This limited observability may lead to ill-posed parameter estimation problems when trying to infer model parameters from given time series data, which is also identified as core failure mode of physics-informed neural networks applied to dynamical systems \cite{Ahmadi_2026_PIML,NEURIPS2021_df438e52}.
\new{Often, these degeneracies are unrecognized, or masked by optimization choices and the explicit integration of symmetry awareness, or structure-informed constraints, are still an open and active research area \cite{burger_learned_2025}.}

The problem of \mjc{determining} whether a given model allows for unique inference of model parameters has been studied extensively in the literature over the past 50 years and is known as Structural Identifiability (SI) \cite{pmlr-v235-balsells-rodas24a}.
If a given model is not structurally identifiable, then multiple parameter configurations will produce identical input-output behaviour of the model. 
It follows that parameters cannot be meaningfully estimated, regardless of the amount and quality of the available data. 
\kb{SI is closely related to the well-known control-theoretic concept of observability.
\new{A dynamical system is called observable, if it is possible to determine its internal state from measurements of the output \cite{hermann1977nonlinear}.
When system parameters are viewed as state variables with zero dynamics, SI analysis may be regarded as a generalized observability problem \cite{villaverde2016structural}.
In the following, we adopt the notion of ``partially observed" as introduced in \cite{shen2017classification}, denoting systems for which only a subset of its state variables (or combinations thereof) can be measured.}
Note the principal difference between the notions of ``partially observed" and ``partially observable" systems, the latter being concerned with the possibility of estimating the internal state from measured system output.}
SI is distinguished from Practical Identifiability (PI), which addresses ambiguities in parameter estimation due to noise and unfavourable sampling times of observations. Structural and practical identifiability are connected to epistemic (systematic) and aleatoric (stochastic) uncertainty \cite{Hora1996AleatoryAE}. 
The discussion about the importance of these uncertainty categories reignited \cite{KIUREGHIAN2009} and distinct handling has gained traction in the machine learning community \cite{Huellermeier2021AleatoricAE}. 
Therefore, SI analysis constitutes an important component for the understanding and reduction of epistemic uncertainty for dynamical systems.

A recent review \cite{Anstett-Collin_2020_a_priori} finds that the discipline of SI analysis can be separated into three main branches:
the Output Equality Approach~\cite{bellman1970structural, pohjanpalo1978system, walter1996identifiability}, 
the Local State Isomorphism Approach~\cite{tunali1987new, vajda1989state, sussmann1976existence}, and 
the Differential Algebra Approach~\cite{diop1991nonlinear, ljung1994global, jain2019priori}. 
Each technique has its respective strengths and limitations. 
In particular, the Output Equality Approach is not generally applicable to non-linear models, but offers an intuitive way for analysing linear models.

In cases where a model is unidentifiable, a range of strategies is commonly used to make parameter inference from data feasible.
A straightforward option is to fix selected parameters so that the remaining set becomes identifiable. 
This approach is simple and preserves the interpretability within the domain-specific context. 
Considerable disadvantages include the necessity for profound mechanistic insight into the model used and a reduced potential for interpretation of the model predictions \cite{WIELAND202160}.
Another option \kb{is} to approximate the behaviour of the unidentifiable model with an alternative \kb{identifiable} model, \kb{which is typically nontrivial}.
Finally, one may attempt to reparametrise an unidentifiable model such that all of the new parameters in the resulting model become identifiable. 
Reparametrisation of a given model often requires a time-consuming manual effort in which practitioners enter a cycle of model construction, quantitative simulation\eco{s} and experimental validation of model predictions.
Recent advances in automatic reparametrisation for dynamical models show great \eco{potential} \cite{Ilmer2021, dong2023differential}. 
Notably, the \emph{AutoRepar} extension~\cite{massonis2021autorepar} for the STRIKE GOLDD SI analysis toolbox~\cite{villaverde2016structural} for MATLAB is capable of semi-automatic reparametrisation for ordinary differential equation models involving rational expressions.

\texttt{AutoRepar} employs a notion of identifiability called Full Input-State-Parameter Observability (FISPO).
It goes beyond establishing parameter identifiability and requires that all states of some auxiliary model are observable.
However, requiring a model to be FISPO is a stronger condition than identifiability alone, and thus \texttt{AutoRepar} is not generally applicable.
It works by detecting and eliminating Lie symmetries that can cause unidentifiability.
However, this approach has two major limitations:
(i) no general method exists to determine the type and number of symmetries in a given model, and
(ii) there is no upper bound on the number of Lie derivative terms required to construct the infinitesimal transformation needed for a suitable reparametrisation \cite{massonis2020}.
In summary, even with semi-automatic reparametrisation tools such as \texttt{AutoRepar}, deriving a reparameterised model that is both identifiable and retains domain-specific interpretability remains challenging. 

As reparameterising a given dynamical model is often highly involved, whereas structural identifiability analysis can frequently be carried out more readily, we propose a model-based framework for time series classification that accounts for the unidentifiability of the underlying dynamical model, referred to as ``\myMethodFull{}" (\myMethod{}).
We employ a model-based time series classification where each individual time series is represented through a Maximum A Posteriori estimate (MAP) of the given dynamical model.
We consider Ordinary Differential Equation (ODE) models in which one or more parameters are unidentifiable. 
Instead of representing individual time series as parameter vectors in the original parameter space, we consider a representation in the space of structurally identifiable parameter combinations. 
Any conventional classification framework acting on vectorial data may subsequently be used to train a classifier in this space.

The contribution of this work is threefold.
Firstly, we propose a novel framework (\myMethod{}) for time series classification that represents time series data as identifiable parameter combinations of an otherwise unidentifiable dynamical system. 
Second, we demonstrate the effectiveness of this framework on several relevant dynamical system models commonly encountered in computational biology. 
In particular, we demonstrate that explicitly accounting for model unidentifiability enables accurate classification even with limited training data.
Finally, we emphasize the importance of SI analysis whenever machine learning is applied in conjunction with parametrised dynamical system models, an aspect that has received limited attention despite its critical role in ensuring successful application.

This paper is organized as follows: 
\autoref{sec:methods} reviews a model-based framework for time series classification and introduces our method of \myMethodFull{} (\myMethod{}). 
\autoref{sec:experiments} 
presents three biologically relevant example models used as test beds for \myMethod{} and outlines the experiments used to evaluate its performance.
\autoref{sec:results} reports the experimental results. 
Finally, \autoref{sec:discussion} and \autoref{sec:conclusion} provide a discussion of the results and their implications.

\section{Methods}\label{sec:methods}

In this section, we present a model-based approach for time series classification based on the incorporation of a given dynamical model in the form of \mjc{a set of parametrised Ordinary Differential Equations} (ODEs). 
To do so, we adopt a formalism in which individual time series observations are represented as Maximum A Posteriori (MAP) estimates.
In addition, we present the details of 
the proposed strategy, namely a \textbf{S}tructural-\textbf{I}dentifiability \textbf{M}apping (\myMethod{}). 
\mjc{The application of a \myMethod{}} is possible whenever the underlying dynamical model is structurally unidentifiable, then structural identifiability analysis can be carried out and explicit expressions \mjc{for} identifiable parameter combinations can be determined. 
This notably includes the class of non-linear ODE models with rational expressions of the states, inputs, and parameters, for which software tools such as \emph{SIAN}\cite{Ilmer2021}, \emph{COMBOS}\cite{meshkat2014COMBOS} and \emph{Structural-Identifiability}\cite{dong2023differential} may be used to automatically determine identifiable model parameter combinations\cite{Rey_Barreiro2023}.

\subsection{Model-based representation for time series data}
In the following, we review the basic notions of Bayesian parameter estimation for dynamical models and adapt \mjc{them} for the purposes of time series classification.
Formulations similar to the one given in this work can be found in \cite{coelho2011bayesian, shen2017classification, linden2022bayesian}.

Let $\{ (\mathcal{Y}^{k},c^{k})\}, k = 1,\ldots,N,$ denote a set of $N$ labelled examples of, potentially multivariate, time series data. 
Here $\mathcal{Y}^{k} = \{ \mathbf{t}^k, \mathbf{Y}^k\}$ consists of a collection of time points $\mathbf{t}^k = \{t_{i}^{k}: i = 1,\ldots, L^{k} \}$ together with a collection of \pt{the corresponding} observations $\mathbf{Y}^k = \{\mathbf{y}_{i}^{k}: i = 1,\ldots, L^{k} \}$ for \mjc{the} time series $k$. Furthermore, $c^{k}$ is the associated class label. 
This formulation allows for \pt{different time series} $\mathcal{Y}^{k}$ to be of different lengths, as indicated by $L^{k}$, and be evaluated at different times, as indicated by $\mathbf{t}^k$. 
However, it is assumed that  all observations have the same dimension, i.e., $\mathbf{y}_{i}^{k} \in \mathbb{R}^{r}$. \emph{The task considered is the prediction of a class label $c$, given a new time series $\mathcal{Y}$ of length $L$}. 
The key idea of this framework is to regard each time series as an instance of a dynamical model from a given model class. 
Time series are considered as partial observations of an underlying dynamical model characterized by a set of Ordinary Differential Equations (ODEs)
\begin{equation}\label{eq:dyn_sys}
    \frac{d \mathbf{x}_{t}}{dt} = f(\mathbf{x}_{t};\boldsymbol{\psi}),
\end{equation}
with $\mathbf{x}_{t} \in \mathbb{R}^{d}$ denoting the state vector at time $t$.
The defining mapping $f$ is parametrized by a vector $\boldsymbol{\psi} = (\boldsymbol{\theta}, \mathbf{x}_{0})$, where \pt{$\boldsymbol{\theta}\in \mathbb{R}^{n}$} is a vector of model parameters and the initial state $\mathbf{x}_{0}$, which may or may not be known.
Observations from the underlying ODE are obtained via the measurement function
\begin{equation}\label{eq:output}
    \mathbf{y}_{i}  = \mathbf{h}(\mathbf{x}_{t_{i}}) + \boldsymbol{\epsilon}_{t_{i}},
\end{equation}
where $\boldsymbol{\epsilon}_{t_{i}}$ is the observation\eco{al} noise at time $t_{i}$. 

For simplicity, it is assumed that the initial condition \mjc{vector} $\mathbf{x}_{0}$ is known and that the observational noise is distributed as $\boldsymbol{\epsilon}_{t_{i}} \sim \mathcal{N}(\mathbf{0}, \mathbf{R})$, i.e. Gaussian with zero mean and covariance matrix $\mathbf{R}$. 
\kb{Should $\mathbf{x}_{0}$ be unknown, it may be treated as an additional unknown system parameter.
If $\mathbf{R}$ is unknown, it may be estimated from the variation of the observed measurements.}
The parameter configuration that is most likely to have produced an observation $\mathcal{Y}$, given a prior $p(\boldsymbol{\theta})$ over the parameters, is the \emph{Maximum A Posterior} (MAP) estimate $\boldsymbol{\theta}_{\text{MAP}}$, which
is the (global) maximum of the posterior distribution
\begin{equation}\label{eq:posterior}
   p(\boldsymbol{\theta} \mid \mathcal{Y}, \mathbf{R}) =  p(\boldsymbol{\theta} \mid \mathbf{Y}, \mathbf{t}, \mathbf{R}) \propto p(\mathbf{Y} \mid \boldsymbol{\theta}, \mathbf{t}, \mathbf{R}) \  p(\boldsymbol{\theta}).
\end{equation}
Under the assumptions made in Eq.\eqref{eq:output}, the likelihood function takes \eco{on} the form
\begin{equation}\label{eq:likelihood}
    p(\mathbf{Y} \mid \boldsymbol{\theta}, \mathbf{t}, \mathbf{R}) = \prod_{i = 1}^{L} \mathcal{N}(\mathbf{y}_{i} \mid \mathbf{x}_{t}(\boldsymbol{\theta}), t_{i}, \mathbf{R}). 
\end{equation}
Finally, for the purposes of this work, we assume that the prior distribution is of the \pt{``bounding box'' form}
\begin{equation}
    p(\boldsymbol{\theta}) = 
    \begin{cases}
        \frac{1}{V(R)}   &\text{if } \boldsymbol{\theta} \in R, \\
        0                &\text{otherwise,}
    \end{cases}
\end{equation}
where $R = [\theta_{1}^{\text{min}}, \theta_{1}^{\text{max}}] \times \ldots \times [\theta_{n}^{\text{min}}, \theta_{n}^{\text{max}}]$ is \eco{the} hyper-rectangle \eco{enclosed by the individual parameter bounds $\theta_i^{\min}, \theta_i^{\max}$} and $V(R)$ \mjc{is} the volume of $R$. 
The set $R$ will be referred to as Region of Interest (ROI).
This prior information essentially restricts the considered region of the parameter space to $R$ but does not provide any additional information\eco{, i.e., is uniform over the region $R$}.
\pt{Interval priors are quite common in biological models, since often only ``physiologically realistic" parameter ranges are known without further probabilistic structure.}
In order to find \eco{the} $\boldsymbol{\theta}_{\text{MAP}}$ associated with a given time series observation, Eq.~\eqref{eq:posterior} is maximized w.r.t. $\boldsymbol{\theta}$, which is equivalent to maximizing Eq.~\eqref{eq:likelihood} subject to $\boldsymbol{\theta} \in R$.

\subsection{Structural-Identifiability Mapping (\myMethod{})}
\begin{figure}[t!]
    \centering
    \includegraphics[width=0.9\linewidth]{
    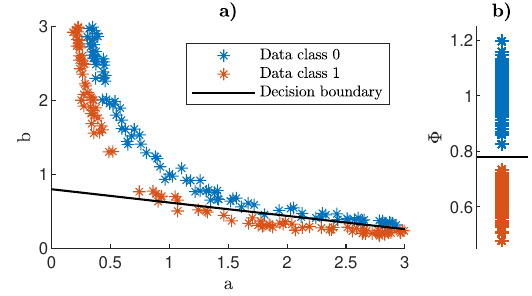}
    \caption{Geometric intuition behind the mechanism of \myMethod{} with data from the toy model. Panel \textbf{a)} depicts a binary classification problem which illustrates how training data can be oriented along manifolds of the form $\Phi = g(a,b) = a b$. Panel \textbf{b)} shows the representation of the same data after applying \myMethod{}. The decision boundary between the two classes becomes simpler and, in this special case, the data even \mjc{become} linearly separable in $\Phi$-space.}
    \label{fig:toy_model_decision_boundary}
\end{figure}

Suppose that the dynamical model given in Eq.~\eqref{eq:dyn_sys} is unidentifiable and that, by means of Structural Identifiability (SI) analysis, it is possible to find a set of identifiable parameter combinations $\boldsymbol{\Phi}$ explicitly characterized by $\boldsymbol{\Phi} = g(\boldsymbol{\theta})$, with $g: \mathbb{R}^{n} \rightarrow \mathbb{R}^{m}$. 
Here, the number of identifiable parameter combinations $m$ is always less than the number of original system parameters $n$, i.e. $m < n$. 
\pt{Consider an equivalence relation on the space of our mechanistic models that identifies models that are behaviourally indistinguishable. 
The equivalence classes of models (parameters) $\mathcal{M}_{\boldsymbol{\Phi}}$ can be then defined as follows:}
\begin{equation}
    \mathcal{M}_{\boldsymbol{\Phi}} = \{ \boldsymbol{\theta} \in \mathbb{R}^{n} \mid \boldsymbol{\Phi} = g(\boldsymbol{\theta}) \}.
\end{equation}
By definition of $g$, any two parameters $\boldsymbol{\theta}_{1}, \boldsymbol{\theta}_{2} \in \mathcal{M}_{\boldsymbol{\Phi}}$ will lead to identical system trajectories of the system in Eq.~\eqref{eq:dyn_sys}, \pt{given identical initial conditions}. 
\pt{We can operate in the factor set.}
Indeed, any level-set of the posterior in Eq.~\eqref{eq:posterior} can be written as a union of sets $\mathcal{M}_{\boldsymbol{\Phi}}$ and maximization of the posterior means to identify the set of equivalence classes associated with the maximal posterior value.
As far as the classification task is concerned, there is no need to resolve the available information beyond the level of equivalence classes.
\autoref{fig:toy_model_decision_boundary} provides some visual intuition on the matter.
We propose to utilize \myMethodFull{} (\myMethod{}) given by $g$ for time series classification as follows:
\begin{enumerate}
    \item Find the model-based representation for each time series by means of a MAP estimate, i.e.
    \begin{equation}
        \mathcal{Y}^{k} \mapsto \boldsymbol{\theta}_{\text{MAP}}^{k},
    \end{equation} 
    \begin{equation}
        \text{with}\quad\boldsymbol{\theta}_{\text{MAP}}^{k} =  \arg\max_{\boldsymbol{\theta}} p(\boldsymbol{\theta} \mid \mathcal{Y}^{k}, \mathbf{R}),
    \end{equation}
    \kb{and} posterior as in Eq.~\eqref{eq:posterior}.
    \item Translate each MAP via $g$ to obtain a representation in the space of identifiable parameter combinations
    \begin{equation}\label{eq:SIM}
        \boldsymbol{\theta}_{\text{MAP}}^{k} \mapsto \boldsymbol{\Phi}^{k} := g(\boldsymbol{\theta}_{\text{MAP}}^{k}).
    \end{equation}
    \item Train a vectorial classifier of choice on the transformed data $\{ \boldsymbol{\Phi}^{k} \}_{k = 1}^{N}$.
\end{enumerate}

\myMethod{} provides a practical approach that can be employed whenever 
structurally identifiable parameter combinations 
can be computed, even in cases where an explicit 
reparametrisation of a 
model in terms of such a set of structurally identifiable combinations is \emph{not} 
possible. 
Therefore, \myMethod{} focuses on the ML task at hand rather than the creation of an all-new dynamical model with more favourable identifiability properties.

From a classification point of view, \myMethod{} can be thought of as having a regularizing influence on the learned decision boundary in \pt{the} $\boldsymbol{\theta}$-space. 
If we were to train the classifier in the space of $\boldsymbol{\theta}$, the decision boundary learned from the data could be such that two values $\boldsymbol{\theta}_{1} \neq \boldsymbol{\theta}_{2}$ with $g(\boldsymbol{\theta}_{1}) = g(\boldsymbol{\theta}_{2})$ become associated with different classes.
This results in undesired behaviour, 
as 
SI analysis shows 
that both values of $\boldsymbol{\theta}$ generate 
identical observable output for the 
dynamical model and, accordingly, should 
be associated with the same class.
On the other hand, training the classifier using \myMethod{}, the learned decision boundary in $\boldsymbol{\theta}$-space becomes the union of pre-images $g^{-1}(\boldsymbol{\Phi})$. 
This guarantees that any two models $\boldsymbol{\theta}_{1}, \boldsymbol{\theta}_{2}$ with $g(\boldsymbol{\theta}_{1}) = g(\boldsymbol{\theta}_{2})$ are always associated with the same class.

\section{Experiments}\label{sec:experiments}
Investigation of \myMethod{} is carried out through three experiments.
Experiment 1 compares learning with a partially observed dynamical system when \myMethod{}, using
a fully observed counterpart of the same dynamical system as a baseline. 
Robustness of \myMethod{} with respect to observational noise is studied in experiment 2.
Experiment 3 addresses the robustness of \myMethod{} with respect to sparsity and irregularity in the time series observations.
These experiments are performed on synthetic data generated by four example systems of increasing complexity, as introduced in \autoref{sec:models}, detailed in \autoref{sec:accuracy}, and summarized in \autoref{tab:overview}.
MATLAB implementations are publicly available on Github\footnote{\url{https://github.com/janis-norden/Structural_Identifiability_Mapping}}.

\begin{table}[t!]
\centering
\caption{Summary of synthetic example models and experiments. FO\ensuremath{\triangleq}fully observed and PO\ensuremath{\triangleq}partially observed.}
\begin{tabularx}{\linewidth}{@{}
l@{ } @{}p{0.38\linewidth}|
@{ }p{0.062\linewidth}
@{ }>{\arraybackslash}p{0.38\linewidth}}
    \hline
    \textbf{System} & \textbf{Characteristics} & \textbf{Exp.
    } & \textbf{Purpose} \\
    \hline
    toy 
    & simplest illustrative model 
    & \\
    CCM2 & simplest model appearing in applications 
    & \multirow[t]{2}{=}{
    1:} 
    & \multirow{2}{=}{compare FO vs. PO vs. PO + SIM approaches} \\
    CCM4 & larger variant of CCM2 
    & \\
    CML & cannot be reparameterised by scaling transformation 
    & 
    2: & {assess impact of observation noise on SIM} \\
    BR & nonlinear model and unknown initial conditions & 
    3: & assess impact of measurement sparsity and irregularity \\
    \hline
\end{tabularx}
\label{tab:overview}
\end{table}

\subsection{Example Models}\label{sec:models}

\subsubsection{\kb{Toy} Model} 
\kb{is a simple artificial model which allows for intuitive visualization of the \myMethod{} due to its $2$-dimensional parameter space (\autoref{fig:toy_model_decision_boundary})}.
The model equations are given by
\begin{align} \label{eq:TM}
    \begin{split}
        \dot{x}(t) &= -ab x(t), \\
        y(t)    &= x(t), 
    \end{split}
\end{align}
where $x \in \mathbb{R}$ is the state variable with $x(0)=1$ known, $t \in [0, 1]$, and $a,b \in \mathbb{R}^{+}$ are the system parameters. 
The parameters are further restricted to lie within the region of interest \eco{(ROI)} $R = [0.1, 3] \times [0.1, 3]$.
From Eq.~\eqref{eq:TM} it can be seen that any parameter configuration $a$ and $b$ such that $\Phi = ab$ is constant, will produce identical system output for a given value of $\Phi$. 

\subsubsection{Catenary compartmental Model (CCM)}
\begin{figure}[t!]
    \centering
    \includegraphics[width = \linewidth]{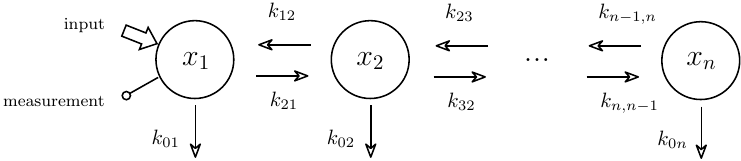}
    \caption{Catenary $n$-compartmental model.}
    \label{fig:CCM}
\end{figure}

\eco{C}ompartmental models are commonly used in modelling pharmacokinetic interactions \cite{jacquez1972, metzler1971, sager2015}. 
The $n$-compartment catenary model (CMM$n$), is a linear model of $n$ compartments which are connected to one another in a \pt{bi-directional chain}.
Only the first compartment is assumed to have an input, \eco{whereas} all compartments are assumed to have leakage (see \autoref{fig:CCM}). 
Substance $x_{i}$ is converted to $x_{i+1}$ and vice versa. 
The model has a total of $3n - 2$ parameters comprising $2(n-1)$ conversion rates and $n$ leakages. 
The coefficients $k_{i,i-1} \geq 0$ and $k_{i-1,i} \geq 0$ describe the conversion rates between $x_{i}$ and $x_{i-1}$, while the coefficients $k_{0i} \geq 0$ govern the leakage. 
The concentration of interacting substances $x_{1},\ldots, x_{n}$ is described by the set of linear ODEs:
\begin{align}\label{eq:CM_dynamics}
    \begin{split}
        \dot{\mathbf{x}}(t) &= K\mathbf{x}(t) + \mathbf{b}u(t)\\
        y(t) &= x_{1}(t),
    \end{split}
\end{align}
where $\mathbf{b} = [1,0,\ldots,0]^{\top}$, $\mathbf{x} = [x_{1},\ldots,x_{n}]^{\top}$ with $\mathbf{x}(0) = \mathbf{x}_{0}$, and $y$ is the system output.
The matrix $K$ is then given by
\begin{equation}
K = \begin{bmatrix}
        k_{11} & k_{12} & 0 & 0 & \ldots & 0 \\
        k_{21} & k_{22} & k_{23} & 0 & \ldots & 0 \\
        0 & k_{32} & k_{33} & k_{34} & \ldots & 0 \\
        0 & 0 & k_{43} & k_{44} & \ldots & 0 \\
        \vdots & \vdots & \vdots & \vdots & \ddots & \vdots \\
        0 & 0 & 0 & 0 & k_{n,n-1} & k_{n,n}
    \end{bmatrix}, \text{ with}
\end{equation}
\begin{equation}
    k_{ii} =
    \begin{cases}
        -k_{01} - k_{21}, & \text{for} \quad i = 1, \\
        -k_{0i} - k_{i+1,i} - k_{i-1,i}  & \text{for} \quad i = 2,3,\ldots, n-1, \\
        -k_{0n} - k_{n-1,n} & \text{for} \quad i = n. \\
    \end{cases}
\end{equation}
Employing the Laplace transform Output Equality approach, Chen~et~al.~\cite{CHEN198559} demonstrated that $2n-1$ structurally identifiable parameter combinations can be found for
\eco{CCM}, namely:
\begin{align}
    \Phi^1_{i} &= k_{ii}, &\text{for} \quad i = 1,2,\ldots, n, \\
    \Phi^2_{j} &= k_{j,j-1} k_{j-1,j}, &\text{for} \quad j = 2,3,\ldots, n.
\end{align}
Using AutoRepar, we were able to reparametrise the CCM2 model to a FISPO model.
The same is not true for the CCM4 model (see Appendix A for details).

We consider the work by Bunte~et~al.~\cite{bunte2018learning} \kb{for experimentation,} \eco{in which} the data from a clinical study concerning the interaction between the metabolites prednisone and prednisolone \eco{is analysed}. 
The authors employed a 3-compartment model \pt{and used a probabilistic mixture of such models} to analyse the data of 12 patients and found that the patients could be stratified into 4 groups.
Their 3-compartment model is equivalent to a CCM \eco{with 2 compartments} with non-zero input and is therefore suitable for the application of a \myMethod{}.
The input in this case is $u(t) = S_{0} k_{\text{abs}} e^{-k_{\text{abs}} t}$, where $S_{0}$ is a fixed amount of prednisone formulation that is ingested and absorbed with rate $k_{\text{abs}}$ into the bloodstream.
The time interval of interest is $t \in [0, 240]$ seconds and the ROI
is $R = [0, 0.1]^{4}$.

To set up a suitable binary classification problem, we use the parametrisation of one of the clusters (C4) found in \cite{bunte2018learning} to represent one \eco{of the} class\eco{es}.
The other class is characterized by the same parametrisation, where the conversion rates $k_{12}$ and $k_{21}$ are $20 \%$ deficient (see \autoref{tab:gt_parameters}).
This model will be referred to as CCM2 and is of particular interest, since it represents a minimal realistic compartmental model for which structurally unidentifiable parameters occur.

We further consider a 4-compartment variant of this model by adding two additional compartments in accordance with the \eco{model schematic} in \autoref{fig:CCM}.
For \eco{the first} class, the excretion and conversion are \eco{the same as} those used for the CCM2.
The second class is characterized by the same parametrisation but now six \eco{out of seven} conversion rates are set to be $50 \%$ deficient.
The \eco{studied} time interval is the same as for CCM2 and the ROI is $R = [0, 0.1]^{10}$.

\subsubsection{Compartmental Model with a Loop (CML)}
\begin{figure}[t!]
    \centering
    \includegraphics[width = 0.6\linewidth]{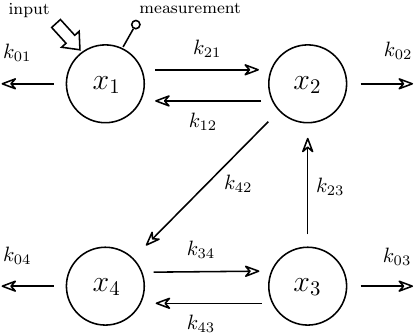}
    \caption{Compartment Model with a Loop (CML).}
    \label{fig:CML}
\end{figure}

To further demonstrate 
\myMethod{} with 
models which cannot be meaningfully reparametrised in a straightforward manner, the Compartmental Model with a Loop (CML) is considered (see \autoref{fig:CML}). 
Similar to the CCM, the CML is a linear compartment model and dynamics Eq.~\eqref{eq:CM_dynamics} apply with coefficient matrix
\begin{equation}
    K = 
    \begin{bmatrix}
        k_{11} & k_{12} &        0 & 0      \\
        k_{21} & k_{22} &   k_{23} & 0      \\
            0  &      0 &   k_{33} & k_{34} \\
            0  & k_{42} &   k_{43} & k_{44}
    \end{bmatrix},
\end{equation}
\begin{align}
    \text{where } \quad
        \notag k_{11} &= -(k_{01} + k_{21}), \\
        \notag k_{22} &= -(k_{02} + k_{12} + k_{42}), \\
        \notag k_{33} &= -(k_{03} + k_{23} + k_{43}), \\
        k_{44} &= -(k_{04} + k_{34}).
\end{align}
and the ROI is given as $R = [0, 0.1]^{10}$. 

Employing the Laplace transform approach, it can be shown that the system has 7 structurally identifiable parameter combinations $\Phi_{1},\ldots, \Phi_{7}$. 
The relations are
\begin{align} \label{eq:CML_SI}
    \begin{split}
        \Phi_{1} &=  k_{12} k_{21}, \\
        \Phi_{3} &=  k_{01} + k_{21}, \\
        \Phi_{4} &=  k_{02} + k_{12} + k_{42}, \\
    \end{split}
    \begin{split}
        \Phi_{2} &=  k_{34} k_{43}, \\
        \Phi_{6} &=  k_{04} + k_{34}\\
        \Phi_{5} &=  k_{03} + k_{23} + k_{43},
    \end{split} \\ 
    &\Phi_{7} =  k_{23} k_{42} k_{34}. \notag
\end{align}
Meshkat~\&~Sullivant~\cite{MESHKAT201446} demonstrate that for this system (Example 6.3 in their work) \kb{no} scaling transformations \kb{exist} which make the resulting reparametrised system identifiable.
We tried AutoRepar with an univariate Ansatz polynomial of degree 2 but could not find any transformations that would make the reparametrised model FISPO. 
Yet, using the same Ansatz polynomial, we were able to find the relations given in Eq.~\eqref{eq:CML_SI} by only looking at parameter identifiability.
Since the CML could not easily be reparametrised, the model is particularly interesting as a test case for \myMethod{}. 

\subsubsection{Batch reactor (BR)}
A classical model \eco{defined} to study microbial growth in a batch reactor which incorporates a Michaelis-Menten type nonlinearity is the following:
\begin{align}
    \begin{split}
        \dot{x}(t) &= \frac{\mu_{m}s(t) x(t) }{K_{s} + s(t)} - K_{d} x(t), \\
        \dot{s}(t) &= -\frac{\mu_{m} s(t) x(t)}{Y(K_{s} + s(t))}, \\
        y(t) &= x(t),
    \end{split} \label{eq:BR_model}
\end{align}
where $x$ is the concentration of microorganisms, $s$ the concentration of growth-limiting substrate, $\mu_{m}$ the maximum reaction velocity, $K_{s}$ the Michaelis-Menten constant, $Y$ the yield coefficient, and $K_{d}$ the decay rate coefficient (see e.g. \cite{button1985kinetics}).
For the present work, the time interval of interest is $t \in [0, 12]$ hours and the ROI is:
\begin{equation}
    R = [0, 10] \times [0, 50] \times [0, 1] \times[0, 5] \times [0, 1] \times [0, 1], 
\end{equation}
where the intervals refer to the allowed ranges of $b_{1}, b_{2}, \mu_{m}, K_{s}, Y$ and $K_{d}$, respectively.

\begin{figure}[t!]
    \centering
    \includegraphics[width=0.9\linewidth]{
    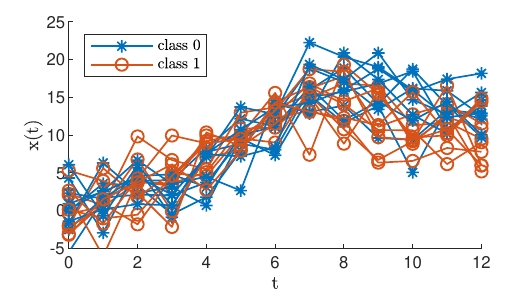}
    \caption{Binary classification task for time series from the batch reactor model. 
    Displayed are 10 time series per class.
    Observational noise simulated is normally distributed with standard deviation $\sigma = 3$.
    }
    \label{fig:BR_example_ts}
\end{figure}

It is assumed that microorganisms $x$ and substrate $s$ are prepared in mixtures for which the concentration can be controlled. 
When the mixtures are put together in the batch reactor, it is assumed that the reaction is very fast, so that the model may be regarded as having impulsive inputs $b_{1}\delta(t)$ for $x$ and $b_{2}\delta(t)$ for $s$.
Equivalently, system Eq.~\eqref{eq:BR_model} may be considered with initial conditions $x(0)=b_{1}$ and $s(0)=b_{2}$.
As demonstrated in \cite{holmberg1982practical}, if both $x$ and $s$ are observed at time $t=0$, then the model is globally structurally identifiable.
However, in \cite{chappell1992structural, evans2000extensions} it was demonstrated that when only $x$ is observed, the model becomes structurally unidentifiable.
In this case, the following combinations of parameters have been found to be structurally identifiable
\begin{align}
    \begin{split}
        \Phi_{1} =  b_{1}, \quad \Phi_{2} &=  \mu_{m}, \quad \Phi_{3} =  K_{d}, \\
        \Phi_{4} &=  b_{2} Y, \quad \Phi_{5} =  \frac{b_{2}}{K_{s}}.
    \end{split}
\end{align}
Realistic configurations of model parameters have been taken from \cite{holmberg1982practical} (cf. Figure 1 in their work).
As a classification task, we consider a scenario in which two reactions are compared which primarily differ in their yield coefficient $Y$.
Class 0 is characterized by a distribution of yield coefficients centred around $Y=0.6$ while class 1 is associated with a distribution that centres around a $20\%$ diminished yield coefficient, i.e., $Y=0.48$.
\autoref{fig:BR_example_ts} illustrates the classification task in the signal space of time series.

\subsection{Experimental setup}\label{sec:accuracy}
In order to test the effectiveness of \myMethod{} approach, a binary classification is implemented based on the Support Vector Machine (SVM) framework. 
For each example system, synthetic time series data corresponding to a binary classification task \mjc{are} created and the SVM classifier is trained using the discussed model-based framework. 
The performance of the resulting classifier is assessed by its generalization error. 
Additionally, the number of support vectors is considered as an indication of the classifier-complexity needed to distinguish the two classes. 
Since the example systems differ in the dimensionality of their parameter spaces, their training and test sets contain differing numbers of training examples $N_{\text{train}}$ and test examples $N_{\text{test}}$.
\eco{For each system}, $N_{\text{train}}$ and $N_{\text{test}}$ are chosen to be sufficiently large as not to be the limiting factor in the assessment of classification performance.

\subsubsection{Experiment 1}
This experiment compares classification performance for three situations: training with the fully observed (FO) dynamical model, training with the partially observed (PO) dynamical model, and training with the partially observed dynamical model together with a \myMethod{} (PO + \myMethod{}).
The synthetic data $\mathbb{D} = \{ (\mathcal{Y}^{k},c^{k}): \quad k = 1,\ldots,N \}$ used for this experiment \mjc{are} generated as follows. 

The ground truth class-conditional distributions associated with classes $c=0$ and $c=1$ are chosen as multivariate normal distributions with known means and covariance matrices:
\begin{equation}
    p(\boldsymbol{\theta} \mid c_{i}) = \mathcal{N}(\boldsymbol{\theta}, \boldsymbol{\mu}_i, \Sigma_{i}), \quad i \in\{0, 1\}.
\end{equation}
The values of $\boldsymbol{\mu}_{i}$ and $\Sigma_{i}$ used for experimentation are specific to the dynamical model under consideration (see \autoref{tab:gt_parameters}).

\begin{table*}[t]
    \centering
    \caption{Ground truth parameter configurations for binary classification tasks of the different models.}
    \begin{tabularx}{0.978\linewidth}{
    l| *{2}{p{0.04\linewidth}} *{3}{p{0.23\linewidth}}
    }
    \hline
    System & $N_{\text{train}}$ & $N_{\text{test}}$ & $\boldsymbol{\mu}_{0}$ & $\boldsymbol{\mu}_{1}$ & $\Sigma_{0}, \Sigma_{1}$  \\
    \hline
    Toy model & 100 & 200 & $(a,b) = (1,1)$ & $(a,b) = 0.9 \cdot (1,1)$ & $10^{-4} I_{2}$ \\
    CCM2 & 100 & 200 & \parbox[t]{\linewidth}{$(k_{01}, k_{02}, k_{12}, k_{21})$\\$ = (0.015,0.015,0.074,0.01)$}  
                     & \parbox[t]{\linewidth}{$(k_{01}, k_{02}, k_{12}, k_{21})$\\$= (0.015,0.015,0.059,0.008)$} & $10^{-7} I_{4}$ \\
    CCM4 & 800 & 1000   & \parbox[t]{\linewidth}{$k_{0i} = 0.015$,\\$(k_{12}, k_{23}, k_{34}, k_{21}, k_{32}, k_{43})=$\\$10^{-2}(7.4,1,7.4,1,7.4,1)$}
                        & \parbox[t]{\linewidth}{$k_{0i} = 0.015$,\\$(k_{12}, k_{23}, k_{34}, k_{21}, k_{32}, k_{43})=$\\$10^{-2}(3.7,0.5,3.7,0.5,3.7,0.5)$}
                        & $10^{-7} I_{10}$ \\
    CML & 800 & 1000 & identical to CCM4 & identical to CCM4 & $10^{-7} I_{10}$ \\
    BR & 200 & 400  & \parbox[t]{\linewidth}{$(b_{1}, b_{2}, \mu_{m}, K_{s}, Y, K_{d})$\\$= (1.25,30,0.5,3,0.6,0.05)$} 
                    & \parbox[t]{\linewidth}{$(b_{1}, b_{2}, \mu_{m}, K_{s}, Y, K_{d})$\\$= (1.25,30,0.5,3,0.48,0.05)$}
                    & \parbox[t]{\linewidth}{$\text{diag}(10^{-2}v),$\\$v=(1, 100, 10^{-2}, 1, 10^{-2}, 10^{-4})$}  \\
    \hline
    \end{tabularx}
    \label{tab:gt_parameters}
\end{table*}

An equal number of example pairs $(\mathcal{Y}^{k},c^{k})$ is generated for each class by \eco{first} drawing $\boldsymbol{\theta}$ from the associated class-conditional distribution $p(\boldsymbol{\theta} \mid c^{k})$ and \eco{subsequently} integrating the dynamical system Eq.~\eqref{eq:dyn_sys} with the drawn $\boldsymbol{\theta}$ on the time interval $t \in [0, t_{\text{end}}]$.
\eco{To obtain $\mathcal{Y}$}, time points are sampled from $[0, t_{\text{end}}]$ and the trajectory of the dynamical system is evaluated at these time points.
This leads to a classification problem with balanced classes.
For the fully observed dynamical model, the system output mapping is assumed to be the identity, i.e. $h(\mathbf{x}) = \mathbf{x}$, and data for each state variable are generated.
For the partially observed model, the system output mapping is the projection onto the first state variable $h(\mathbf{x}) = x_{1}$.

For experiment 1, a regular time grid on which the data are generated is chosen to densely cover the time interval of interest\eco{: $t_{\text{dense}}$}.
The collection of observations $\mathbf{Y}^k$ is then obtained by evaluating the resulting trajectory $\mathbf{x}(t;\boldsymbol{\theta})$ on the time grid and adding observational noise, which is assumed to be Gaussian, i.e.,
\begin{equation}\label{eq:output_scalar}
    \mathbf{y}_{i}^{k} = \mathbf{x}(t_{i}^{k}) + \boldsymbol{\epsilon},
\end{equation}
where $\boldsymbol{\epsilon} \sim \mathcal{N}(\mathbf{0}, \mathbf{R})$ and $\mathbf{R}$ is known.
The observable output $\mathbf{y}_{i}^{k}$ of the FO model has a different dimension than the one of the PO model.
Therefore, we distinguish between $\mathbf{R}=\mathbf{R}_{FO}$ and $\mathbf{R}=\mathbf{R}_{PO}$.
Details about \eco{what} time grid is used for \eco{each} model, as well as the matrices $\mathbf{R}$, are reported in Table~\ref{tab:exp_conditions}.

\eco{Once} the set of labelled time series data $\mathbb{D}$ \eco{is obtained}, this set \kb{is transformed} into a set of labelled Maximum \emph{A Posteriori} inferential model parameter estimates
$\mathbb{D}_{\text{MAP}} = \{ \boldsymbol{(\theta}^k_{\text{MAP}}, c^{k}) \}_k$.
Note that since we have chosen a flat prior over $R$, $\boldsymbol{\theta}^k_{\text{MAP}}$ will be a maximum likelihood estimate constrained to $R$:
\begin{equation}\label{eq:argmax}
    \boldsymbol{\theta}_{\text{MAP}}^{k} = \arg\max_{\boldsymbol{\theta} \in R} \Big\lbrace \log(p(\mathbf{Y}^{k} \mid \boldsymbol{\theta}, \mathbf{t}^{k}; \mathbf{R})) \Big\rbrace.
\end{equation}
\eco{We remark that} the argmax operation does not determine $\boldsymbol{\theta}_{\text{MAP}}^{k}$ uniquely, due to structural unidentifiability \eco{of the system}. 
The problem given in Eq.~\eqref{eq:argmax} is solved using MATLAB's {\tt simulannealbnd} function which can be used for constrained optimization. 
The outcome of the data \pt{transformation} process is \pt{the set $\mathbb{D}_{\text{MAP}}$ and its \myMethod{} counterpart ${\mathbb{D}_{\text{\myMethod{}}} = \{ (\boldsymbol{\Phi}^{k}, c^{k}) \}_k}$,} where each $\boldsymbol{\Phi}^{k}$ is obtained as described in Eq.~\eqref{eq:SIM}.
For the fully observed dynamical model, only $\mathbb{D}_{\text{MAP}}$ is generated, whereas for the partially observed (and thus unidentifiable) model, both $\mathbb{D}_{\text{MAP}}$ and  $\mathbb{D}_{\text{\myMethod{}}}$ are produced.

Once the sets $\mathbb{D}_{\text{MAP}}$ and $\mathbb{D}_{\text{\myMethod{}}}$ have been created, Support Vector Machine (SVM) classifiers are trained to learn the binary classification rule. 
\pt{A separate hold-out test data set (never used for training) is employed to assess the generalisation performance.}
The number of training and test examples produced is also reported in Table~\ref{tab:gt_parameters}.
As \eco{an} increasing \eco{number} of training \eco{examples} are made available, the generalization error and the relative number of support vectors are \pt{recorded} as a function of the number of training examples per class.
For each number of available training examples per class, the classifier is trained on 30 randomly sub-sampled datasets, and the mean and standard deviation of the generalization error 
and relative number of support vectors are reported as a function of the number of training examples.
Training the classifier for 30 independent trials permits the capture of the variability in classification performance for a given number of training examples while keeping the runtime of the experiments feasibly low.
 
For SVM training we use MATLAB's {\tt fitcsvm} with a Gaussian kernel.
The kernel scale and the hyper-parameter governing the penalization of misclassification ({\tt BoxConstraint}) are selected by means of $10$-fold cross-validation for each round of classifier training.

\subsubsection{Experiment 2}
\kb{This experiment is designed to study the robustness of \myMethod{} with respect to observational noise.
For this purpose only the PO model and the PO model + \myMethod{} are compared.
The overall setup mirrors that of experiment $1$ with a few key differences.
\eco{First, t}he number of training examples per class is kept constant, while the amount of observational noise is varied. 
We achieve this by setting $\mathbf{R} = \sigma^2 I$, where $\sigma$ is adjusted to correspond to different levels of signal-to-noise ratio (SNR). 
Since all time series in experiment 2 are univariate, we assume a given time series of outputs
\begin{equation}
    y_{i}  = h(\mathbf{x}_{t_{i}}) + \epsilon_{t_{i}}, \quad i = 1, \ldots, L,
\end{equation}
with $\epsilon_{t_{i}} \sim \mathcal{N}(0, \sigma^2)$. 
Furthermore, the power of the observed signal is defined as:
\begin{equation}
    P_\text{signal} = \frac{1}{t_{L} - t_{1}} \int_{t_{1}}^{t_{L}} h(\tau)^2 d\tau,
\end{equation}
with the power of the noise 
given by $P_\text{noise} = \sigma^2$ \cite{papoulis1965random}.
The SNR is the ratio: $SNR = P_\text{signal}/P_\text{noise}$, where the range $ 1 \leq \text{SNR} \leq 10^3$ is considered.
Changes in the observational noise are applied to both training and test data.
For each value of $\sigma$, the classifier is 
evaluated on 50 randomly sub-sampled datasets.
The mean and standard deviation of the generalization error, as well as relative number of support vectors, are reported. 
Additionally, the following three quantities are computed for each example model. 
1) $\Delta \epsilon^{\ast}$ (the maximum difference in mean generalization error), 
2) $\text{SNR}^{\ast}$ (the SNR at which that maximum occurs), and 
3) $\langle \Delta \epsilon \rangle$ denoting 
the average of the difference between the generalization error curves obtained for the PO model and the PO model + \myMethod{}.}

\subsubsection{Experiment 3}
This experiment is designed to study the robustness of \myMethod{} with respect to sparsity and irregularity \eco{of} the time series data.
Again, the overall setup is identical to that of experiment 1.
In contrast to \eco{experiment} 2, the observational noise is fixed.
\kb{Time series are generated on three different types of time grids: 
A dense grid $t_{\text{dense}}$, which corresponds to frequent and regular measurements; a sparse grid $t_{\text{sparse}}$, which is regular like $t_{\text{dense}}$ but only contains $40 \%$ of the points, and 
irregular grids $t^{k}_{\text{irr}}$ containing 
sparse and irregular measurements 
different for every observation $k$.}  
Each irregular time grid contains between $25\%$ and $40\%$ of the number of points in $t_{\text{dense}}$.
Time points in the irregular grid are sampled uniformly at random between the first and last time points in $t_{\text{dense}}$.
Notably, all time grids contain $t=0$.
The choices for $t_{\text{dense}}$ for the different example models are reported in Table~\ref{tab:gt_parameters}. 
\kb{The configurations of $t_{\text{sparse}}$ and $t^{k}_{\text{irr}}$ follow from the choices of $t_{\text{dense}}$.}
In addition to the PO model and PO model + \myMethod{}, we also considered classification by means of a Multi-Layer Perceptron (MLP), which is not model-based and required some additional strategy to deal with the time series on the irregular grids.
To this end we used interpolation by means of Linear Regression and Gaussian Process models.
While for the dense and sparse grids, the MLP methods tend to perform better, we observed catastrophic deterioration of the MLP performance on the irregular grids for most of the example models, leading us to not pursue this approach any further.
Finally, for experiment $3$, \eco{the} mean and standard deviation of the generalization error are reported for the three differ\eco{ent types of} time grids.

\begin{table}[t!]
    \centering
    \caption{\kb{Experimental configurations for parameters related to time grids and observational noise.}}
    \begin{tabular}{lcrc}
        \hline
        System & $t_{\text{dense}}$ & $\mathbf{R}_{FO}$ & $\mathbf{R}_{PO}$ \\
        \hline
        Toy model & $0,0.1,\ldots, 1.0$  & 0.01                   & -       \\
        CCM2      & $0, 10,\ldots, 240$ & $10^2 \mathbf{I}_{2}$  & $10^2$  \\
        CCM4      & $0, 10,\ldots, 240$ & $10^2 \mathbf{I}_{4}$  & $10^2$  \\
         CML      & $0, 10,\ldots, 240$ & $10^2 \mathbf{I}_{4}$  & $10^2$  \\
         BR       & $0, 1, \ldots, 12$  & $\mathbf{I}_{2}$       & $1$     \\    
         \hline
    \end{tabular}
    \label{tab:exp_conditions}
\end{table}

\kb{
\subsubsection{Experiment 4}
To validate \myMethod{} on a real-world dataset we use the prednisone conversion model and dataset studied in \cite{bunte2018learning}.
Time series taken from 12 participants are found to group into three classes: \emph{slow absorbers}, \emph{medium absorbers} and \emph{fast absorbers} of prednisone.
The dataset is imbalanced: there are 2 slow absorbers, 5 medium absorbers and 5 fast absorbers.
The time series have varying number of measurements, ranging from 5 to 12 observed time points.
To increase the degree of data sparsity every 2nd measurement is removed for demonstration.
For the multi-class classification problem we deploy an Error Correcting Output Codes (ECOC) model with Support Vector Machines (SVM) as binary learners\footnote{Implemented in MATLAB's \texttt{fitcecoc}.}.
We use leave-one-out cross-validation to estimate the out-of-sample performance and employ the SVMs hyper-parameter tuning as described in experiment 1. 
We compare the PO model with and without \myMethod{} and summarise 
results in \autoref{fig:exp4_pred}.
}

\section{Results}\label{sec:results}
In the following, the results of experiments 1, 2 and 3 are presented in detail for the batch reactor model example.
The results for all other example models are qualitatively similar and 
summarized in 
\autoref{tab:experiment_1}, \ref{tab:experiment_2} and \ref{tab:experiment_3}.
The results for the other example models are presented in detail in Appendix D. 

\subsection{Experiment 1}

\begin{figure}[t!]
    \centering
    \includegraphics[width=0.9\linewidth]{
    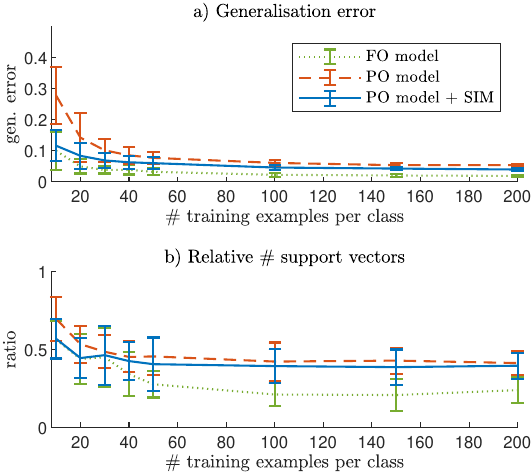}
    \caption{Experiment 1 showing improved classification with partially observed batch reactor model due to \myMethod{}. 
    Displayed are learning curves obtained from classifier training based on the fully observed (FO) dynamical model (dotted green) and the partially observed (PO) dynamical model, with and without application of \myMethod{} (marked with solid blue and dashed orange curves, respectively).
    The training and test data used were generated on the dense time grid $t_{\text{dense}}$ with fixed observational noise $\sigma = 1$ on each observed component.
    }
    \label{fig:exp1_BR}
\end{figure}

\begin{figure}[t!]
    \centering
    \includegraphics[width=0.9\linewidth]{
    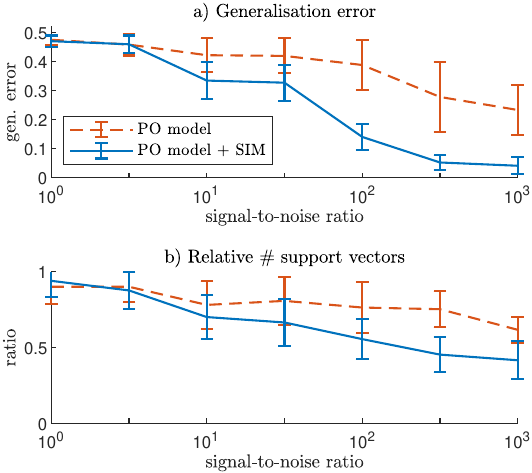}
    \caption{\kb{Experiment 2 for the partially observed batch reactor model showing that \myMethod{} is robust to observational noise.
    Displayed are generalization error and relative number of support vectors, each as a function of the observational noise.
    Classification performance is compared for the partially observed (PO) dynamical model, with and without application of \myMethod{} (marked with solid blue and dashed orange curves, respectively).
    Training and test data are generated on the dense time grid $t_{\text{dense}}$ with $N_{\text{train}} = 20$ and $N_{\text{test}} = 400$.}
    }
    \label{fig:exp2_BR}
\end{figure}

\begin{figure*}[t!]
    \centering
    \includegraphics[width=0.9\linewidth]{
    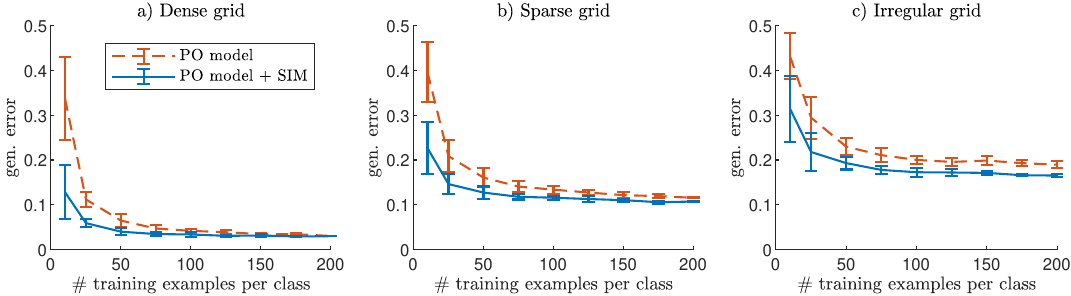
    }
    \caption{Experiment 3 for the partially observed batch reactor model showing \myMethod{} is robust to changes in regularity and sparsity of the observed time series data.
    Displayed are the learning curves obtained from classifier training for the partially observed (PO) dynamical model, with and without application of \myMethod{} (marked with solid blue and dashed orange curves, respectively).
    The training and test data used were generated on the three different time grids $t_{\text{dense}}, t_{\text{sparse}}$ and $t_{\text{irr}}$, displayed in the left, middle and right panel, respectively.
    Observational noise is fixed at $\sigma = 1$ on each observed component.
    }
    \label{fig:exp3_BR}
\end{figure*}

\begin{table*}[ht!]
    \centering
    \caption{
    \kb{Summary of experiment 1 comparing classification with the fully observed (FO) model, partially observed (PO) model and partially observed model with \myMethod{} (PO + \myMethod{}).
    Mean generalization errors (and standard deviations), 
    evaluated at the lowest number of training examples, are shown for all example systems.}
    }
        \begin{tabularx}{0.901\linewidth}{
        l|*{2}{p{0.04\linewidth}}@{ }|*{3}{p{0.095\linewidth}}|*{3}{>{\arraybackslash}p{0.095\linewidth}}|
        } 
            \cline{2-9} 
            & \multicolumn{2}{c|}{Examples} & \multicolumn{3}{c|}{Generalization error at $N_{\min}$} & \multicolumn{3}{c|}{Generalization error at $N_{\max}$}  \\
                System & $N_{\min}$ & $N_{\max}$ & \multicolumn{1}{c}{FO} & \multicolumn{1}{c}{PO} & \multicolumn{1}{c|}{PO + SIM} 
                & \multicolumn{1}{c}{FO} & \multicolumn{1}{c}{PO} & \multicolumn{1}{c|}{PO + SIM} \\
            \hline
            CCM2 & 10 & 100 & .0236 (.0304) & .0734 (.0748) & .0336 (.0230) & .0004 (.0009) & .0042 (.0012) & .0022 (.0016) \\
            CCM4 & 10 & 400 & .0568 (.0430) & .3326 (.0484) & .1855 (.0670) & .0007 (.0007) & .0174 (.0040) & .0133 (.0018) \\
            CML  & 10 & 400 & .0707 (.0645) & .3705 (.0516) & .1923 (.0648) & .0003 (.0005) & .0112 (.0012) & .0098 (.0010) \\
            BR   & 10 & 200 & .0995 (.0605) & .2774 (.0931) & .1158 (.0493) & .0187 (.0022) & .0534 (.0024) & .0392 (.0033) \\
            \hline
        \end{tabularx}
    \label{tab:experiment_1}
\end{table*}

\begin{table}[ht!]
    \centering
    \caption{
    \kb{Summary of experiment 2.
    Maximum ($\Delta \epsilon^{\ast}$) and mean ($\langle \Delta \epsilon \rangle$) difference in generalization error due to SIM.
    The $\text{SNR}$ at which $\Delta \epsilon^{\ast}$ occurs is $\text{SNR}^{\ast}$. Values of $\text{SNR}^{\ast}$ are 
    in $\log$-scale.}
    }
    \begin{tabular}{ 
    l*{4}{p{0.08\linewidth}}
    }
    \hline
        System & $N_\text{train}$ & $\text{SNR}^*$ & $\Delta \epsilon^{\ast}$ & $\langle \Delta \epsilon \rangle$ \\
        \hline
        Toy model   & 10 & 2.5 & .16 & .07 \\
        CCM2        & 10 & 2.5 & .03 & .02 \\
        CCM4        & 10 & 4.0 & .20 & .09 \\
        CML         & 10 & 3.5 & .22 & .09 \\
        BR          & 10 & 2.0 & .23 & .13 \\
        \hline
    \end{tabular}
    \label{tab:experiment_2}
\end{table}

\renewcommand{\arraystretch}{1.155}
\def\myColW{0.114} 
\begin{table}[ht!]
    \centering
    \caption{\kb{Summary of experiment 3: 
    Mean generalization errors (stds) of \myMethod{} with dense, sparse and irregular observation grids and 
    $N_{\min}=10$ training samples for each model.}}
    \begin{tabularx}{\linewidth}{@{}p{0.035\linewidth}@{ }l|*{5}{>{\centering\arraybackslash}p{\myColW\linewidth}}}
        \hline
        \multicolumn{2}{c|}{\diagbox[width=1.7cm,height=0.6cm]{\scriptsize Setting}{\scriptsize System} 
        } 
        & Toy & CCM2 & CCM4 & CML & BR \\
        \hline
        \multirow{2}{*}{\rotatebox{90}{$t_{\text{dense}}$}}
        & PO     & .03 (.04) & .05 (.04) & .33 (.04) & .33 (.05) & .34 (.09) \\
        & PO+SIM & .00 (.00) & .02 (.01) & .16 (.07) & .18 (.05) & .13 (.06) \\ \hline
        \multirow{2}{*}{\rotatebox{90}{$t_{\text{sparse}}$}}
        & PO     & .05 (.04) & .17 (.06) & .41 (.03) & .42 (.03) & .40 (.07) \\
        & PO+SIM & .00 (.00) & .09 (.03) & .27 (.05) & .27 (.06) & .23 (.06) \\ \hline
        \multirow{2}{*}{\rotatebox{90}{$t_{\text{irr}}$}}
        & PO     & .06 (.06) & .19 (.07) & .40 (.03) & .42 (.03) & .43 (.05) \\
        & PO+SIM & .00 (.00) & .17 (.04) & .25 (.06) & .28 (.04) & .31 (.07) \\
        \hline
    \end{tabularx}
    \label{tab:experiment_3}
\end{table}
\renewcommand{\arraystretch}{1}
The outcomes of experiment 1 for the batch reactor model are summarized in \autoref{fig:exp1_BR}.
Comparing the training outcomes of the fully observed (FO) dynamical model \eco{to} the partially observed (PO) dynamical model\eco{,} the results are not surprising.
The training data obtained for the FO model are a super-set of the data available for the PO model.
One would therefore expect that the classifier training with the FO model is more successful than training with the PO model.
This is indeed reflected in \autoref{fig:exp1_BR}. 
For any amount of training data available, the FO curve for the generalisation error lies significantly below the PO curve.
The same is true for the relative number of support vectors.
As expected, using training data which include observations from all compartments, the problem of structural identifiability does not arise and it is possible to achieve better classification performance by fitting models of relatively low complexity.

The outcomes become more interesting when comparing the \eco{performance} of the FO and PO models to \eco{those} for the PO model where \myMethod{} was applied. 
Considering the generalization error, it is clear that the PO model + \myMethod{} outperforms the PO model \pt{consistently} when the number of training examples is less than 50.
Subsequently, the PO model and PO model + \myMethod{} reach comparable levels of generalisation error.
Neither the PO model nor the PO model + \myMethod{} quite reach the performance of the FO model.
A similar situation can be observed for the relative number of support vectors.
Up to 50 training examples, the mean curve for the PO model + \myMethod{} is very similar to the mean curve for the FO model.
After 50 training examples, the mean curve for the PO model + \myMethod{} becomes evermore  similar to that for the PO model.
The FO, PO and PO + \myMethod{} curves obtained for the number of support vectors in \autoref{fig:exp1_BR} are overall very similar to one another (when accounting for the observed standard deviations) and the effect of \myMethod{} is less clearly visible.
However, in the low data regime up to 50 examples, \myMethod{} turns the classification problem into one with a less complex decision boundary associated with fewer support vectors and reduced generalisation error.

The outcomes of experiment 1 for the other models are summarized in \autoref{tab:experiment_1} and provide a similar picture.
When comparing the generalization errors at the maximal number of training examples $N_{\max}$, the values for the PO model and the PO model + \myMethod{} are typically very similar with the values of the FO model, being significantly lower.
However, at the minimal number of training examples, $N_{\min}$, the PO model is clearly outperformed by the PO model + \myMethod{}, which, in turn, is clearly outperformed by the FO model.

The conclusion to be drawn is clear: for densely sampled time series data and with relatively low observational noise present, \myMethod{} approach significantly reduces the complexity of the classification problem.
Notably, when relatively little training data are available, the classification performance is remarkably close to the performance that would be attained if the underlying dynamical model was fully observed.
Utilizing the information from the structural identifiability analysis therefore exhibits a good alternative, when certain measurements are unobtainable,
for improving machine learning performance, in particular when training data are limited.

\subsection{Experiment 2}

The outcomes of experiment 2 for the batch reactor model are summarized in \autoref{fig:exp2_BR}.
Several trends emerge from the results. 
As the SNR increases, both the generalization error and relative number of support vectors decrease, as the classification problem becomes easier with less noise.
When the SNR approaches 1 distinguishing the classes becomes more and more difficult and the generalization error approaches $0.5$ due to the signal degradation.
Nevertheless, for a wide range of SNR values, the generalization error is significantly reduced by utilizing \myMethod{}.
Similarly, using \myMethod{} requires fewer support vectors with higher SNR, and more training samples near the boundary become assigned as the SNR decreases. 
\autoref{tab:experiment_2} summarizes the experimental results for the remaining example models. 
The maximal error difference $\Delta \epsilon^{\ast}$ occurs at high SNR values, as expected, since structural identifiability (in contrast to practical identifiability) describes an inherent model property, that is independent of observational noise, making the \myMethod{} most effective at zero noise ($\text{SNR} \rightarrow \infty$).
The mean difference in the generalization error $\langle \Delta \epsilon \rangle$ is positive for all example models, indicating a 
reduction in error due to \myMethod{} across various noise levels. 
In conclusion, experiment 2 shows that \myMethod{} improves 
classification performance across 
a wide range of observational noise levels, with 
the benefits growing 
as the difficulty increases.

\subsection{Experiment 3}
The outcomes of experiment 3 for the batch reactor model are summarized in \autoref{fig:exp3_BR}. 
It is to be noted that the \eco{presentation} of \eco{the} results for experiment 3 differ from those for experiments 1 and 2 in that 
for this experiment we plot \emph{only} the generalization error. 
The learning curves are qualitatively the same as those presented in \autoref{fig:exp1_BR} and the general trend is again clearly visible. 
Using time series data with observations in $t_{\text{dense}}$ yields better generalization errors than training with data that are generated in $t_{\text{sparse}}$.
Similarly, training with data generated in $t_{\text{sparse}}$ yields better overall results than training on data that are generated in $t_{\text{irr}}$.
This is reasonable, since the data in $t_{\text{dense}}$ simply contain more information than those in $t_{\text{sparse}}$ and $t_{\text{irr}}$.
The effect of \myMethod{} appears to be robust with respect to the time grid used:
for each time grid the application of \myMethod{} yields reduced generalization errors.
Since the difference between the PO model and the PO model + \myMethod{} are most pronounced for relatively small amounts of training data, \autoref{tab:experiment_3} summarizes the outcomes of experiment 3 for all example models at the minimal number of training examples $N_{\min}$.
In all cases, the application of \myMethod{} leads to a reduction in average generalization error.

\begin{figure*}[ht!]
\centering
\includegraphics[width=0.9\textwidth]{
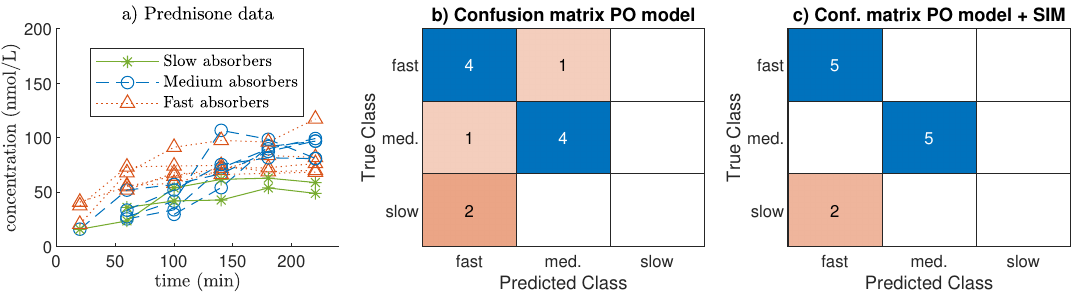}
\caption{\kb{
Experiment 4 validates \myMethod{} on a three-class problem of prednisone conversion with classification measurements shown in panel a). 
The confusion matrix resulting from leave-one-out cross-validation based on the PO model without and with \myMethod{} are depicted in panel b (error $0.33$) and c (error $0.17$).}
}
\label{fig:exp4_pred}
\end{figure*}

\kb{
\subsection{Experiment 4}
As shown in \autoref{fig:exp4_pred} \myMethod{} clearly improves the classification performance of the medium and fast absorbers: 
two examples which are misclassified under the PO model are correctly classified when \myMethod{} is used. 
The 2 slow absorbers cannot be retrieved correctly due to the class imbalance.
}

\section{Discussion}\label{sec:discussion}

\new{The computational complexity of \myMethod{} breaks down into three components: the SI analysis, MAP estimation and machine learning training.
Each of these may be considered independent and the exact choice of algorithm to tackle each step should be tailored to the application at hand.
A standard training pipeline for a model-based approach would involve finding suitable representations of data in the model space (MAP estimation in this paper) and subsequent training of the classifier of choice.
The preceding \myMethod{} analysis is performed once, but its complexity strongly depends on the problem at hand.
Due to these factors and involvement of symbolic manipulation, the computational complexity of SI analysis methods is generally difficult to determine. 
Consequently, existing approaches are primarily assessed through empirical comparisons, as illustrated in the recent benchmarking study \cite{Rey_Barreiro2023} covering a range of SI analysis methods.}

It was observed that the \myMethod{} approach is most effective for relatively small amounts of training data. 
This is because \myMethod{} removes redundancies in the space of the original model parameters and thus makes the decision boundary in the space of identifiable parameter combinations simpler (cf. \autoref{fig:toy_model_decision_boundary}).
As the amount of available data increases, the effect of \myMethod{} is diminished since the additional data now suffice to resolve the class-membership distribution in the space of the original parameters. 
This means that \myMethod{} has a regularizing effect on the classifier training and is particularly useful whenever there are relatively few data available, which is common in biomedical applications. 

Deep Learning learning methods have been applied to the problems of time series analysis and classification with great success.
However, with \mjc{the} large number of weights to be trained, deep networks have a tendency to over-fit and effective regularization becomes a strict necessity when working in the small-data regime. 
Recently, Physics-informed Neural Networks (PINN) have been introduced which regularize the training process by the incorporation of any physical laws in the form of ODEs and/or partial differential equations (PDE) \cite{raissi2019pinn}.
In this context, PINN can also be used for parameter estimation for ODEs (as a special case of PDEs).
However, if a PINN were to be set up incorporating an ODE with unidentifiable parameters, \mjc{then any form} of parameter estimation would again become meaningless.
A thorough Structural Identifiability analysis of the underlying dynamical model is therefore strongly recommended when employing a PINN for parameter estimation.

\eco{In any situation involving high-stakes decision making,} interpretability is of critical importance.
A recent review, \eco{in which} 9 state-of-the-art deep learning methods for time series classification \eco{are compared}, found that only 2 out of the 9 methods studied address the issue of interpreting the decision taken by the neural network~\cite{ismail2019deep}.
Using \myMethod{}, even though a given classifier is trained on data in the space of identifiable parameter combinations, the learned decision boundary can be recovered in the space of the original model parameters.
This makes the learned decision boundary interpretable for domain experts and increases trust in the trained model.
Another example in which insight is generated from an unidentifiable model in a similar manner can be found in Bunte~et~al.~\cite{bunte2018learning}. 
\myMethod{} not only improves classification performance but also preserves interpretability of the model-based approach.

There are a number of limitations to be considered when applying the \myMethod{} approach.
\eco{For one}, the extent to which the existence of \pt{non-trivial output-equivalent manifolds of models} actually hampers classification performance is hard to predict a priori.
Depending on the optimization scheme employed to maximize the log-likelihood function, and depending on the dynamical system \mjc{in question}, performance degradation may be more or less severe, making the effectiveness of \myMethod{} situation-dependent.
Moreover, in \cite{shen2017classification}, the authors point out that working with point estimates (like MAP) to represent time series data in the parameter space of a given dynamical model comes with inherent difficulties because such estimates do not quantify the uncertainty for models \textit{around} these estimates.
As an alternative, the authors propose a fully Bayesian approach and represent each time series observation as a posterior distribution over the entire model parameter space. 

\kb{This paper extends beyond previous work \cite{shen2017classification,bunte2018learning} by incorporating structural identifiability (SI) analysis even for structurally non-identifiable (SNI) models (see \autoref{tab:related_work}). 
While we build on the ``learning in the model space" (LiMS) framework in \cite{shen2017classification}, structural identifiability is not addressed there.
The model-based clustering approach in \cite{bunte2018learning} includes SI analysis, but the model is structurally \emph{locally} identifiable with finitely many solutions, that allowed the exclusion of biologically unfeasible cases. 
In contrast, the present work makes use of SI analysis regardless of whether the model is locally identifiable, enabling us to handle structurally non-identifiable (SNI) models and thus go beyond the scope of \cite{shen2017classification} and \cite{bunte2018learning}.}
\begin{table}[ht!]
    \centering
    \caption{
    SI analysis and 
    SNI models in related works.}
    \begin{tabular}{>{\scriptsize}l>{\scriptsize}c>{\scriptsize}c>{\scriptsize}c}
    \diagbox[width=2cm,height=0.5cm]{\scriptsize Content}{\scriptsize Paper}
    & Shen et al. \cite{shen2017classification} & Bunte et al. \cite{bunte2018learning} & present work
                    \\\hline
    SI analysis     & \ding{55} & \ding{51}    & \ding{51} \\
    Models are SNI  & \ding{55} & \ding{55}    & \ding{51} \\
    \end{tabular}
    \label{tab:related_work}
\end{table}

\section{Conclusion}\label{sec:conclusion}

Model-based approaches for time series classification can be effectively utilized when a model of the underlying dynamical process is available~\cite{shen2017classification}. 
Using structural identifiability (SI) analysis, structurally identifiable parameter combinations of the dynamical model can be obtained. 
Individual time series observations may then be represented as point estimates in the original parameter space or in the space of structurally identifiable parameter combinations. 
We introduced a strategy dubbed \textbf{S}tructural-\textbf{I}dentifiability \textbf{M}apping (\myMethod{}) and demonstrated that \myMethod{} improves classification performance for time series data when taking a model-based approach and the underlying dynamical model is structurally unidentifiable.

It has been shown on a set of relevant example systems that classification performance is significantly improved when learning with data represented in the space of structurally identifiable parameter combinations. 
The increase in performance also persists when time series data of varying quality \mjc{are} \eco{produced}: for all types of time grids (dense, sparse and irregular) as well as for \eco{varying levels of} the observation\eco{al} noise introduced, learning in the space of structurally identifiable parameter combinations outperforms learning in the original parameter space.
This work demonstrates incorporating SI analysis directly into the learning process for classification.
\myMethod{} is straightforward and can be applied whenever a SI analysis can be carried out. 
An explicit reparametrisation of a given dynamical model in terms of fewer, structurally identifiable parameters is not needed in order to benefit from SI analysis. 
This is especially important when explicit expressions for structurally identifiable parameter combinations are available, but suitable model reparametrizations are not possible.

Finally, outcomes of the learning process stay interpretable: while interpretation in the space of structurally identifiable parameter combinations is not straightforward, any insight in this space may be translated back to the space of the original model parameters $g^{-1}(\Phi)$ \pt{which, in turn, are meaningful in the domain-specific context, for example stimulating mathematical modellers to formulate further viable constraints on system configurations,} \new{or simulate interventions to find the most effective or least invasive options.}

\ifCLASSOPTIONcaptionsoff
  \newpage
\fi

\bibliographystyle{ieeetr}
\bibliography{References.bib}

\appendix


\section{Possible Reparametrisation of Example Models}\label{app:repar}

\subsection{FISPO reparametrisation of CCM2 using AutoRepar}
Using AutoRepar, the catenary compartment model with $n=2$ and without input, i.e.
\begin{align}
    \begin{split}
        \dot{\mathbf{x}}(t) &= K\mathbf{x}(t)\\
        \mathbf{y}(t) &= x_{1}(t),
    \end{split}
\end{align}
where $\mathbf{b} = [1,0]^{\top}$, $\mathbf{x} = [x_{1},x_{2}]^{\top}$ with $\mathbf{x}(0) =  [1,0]^{\top}$, and 
\begin{equation}
K = \begin{bmatrix}
        -(k_{01}+k_{21}) & k_{12} \\
        k_{21} & -(k_{02}+k_{12}) 
    \end{bmatrix},
\end{equation}
can be reparametrised to give
\begin{align*}
    \dot{\tilde{x}}_{1} &= -(\tilde{k}_{01} + \tilde{k}_{21}) \tilde{x}_{1} + \tilde{x}_{2}\\
    \dot{\tilde{x}}_{2} &= \tilde{k}_{21} x_{1} + 2 \tilde{x}_{2}, \\
    y &= \tilde{x}_{1},
\end{align*}
using the transformations
\begin{align*}
    \tilde{x}_{1} &= x_{1}, \\
    \tilde{x}_{2} &= (-2 + k_{02} + k_{12}) x_{1} + k_{12} x_{2}, \\
    \tilde{k}_{01} &= 2 - k_{12} + k_{01} (-1 + k_{02} + k_{12}) \\
    &+ k_{02} (-1 + k_{21}) - k_{21}, \\
    \tilde{k}_{21} &= -k_{01} (-2 + k_{02} + k_{12}) - k_{02} (-2 + k_{21}) \\
    &+ 2 (-2 + k_{12} + k_{21}).
\end{align*}
Note that AutoRepar suggests transformations which make a given model Fully-Input-State-Parameter-Observable (FISPO). In this case the reparametrised version of the CCM2 model ends up having 2 parameters ($\tilde{k}_{01}$ and $\tilde{k}_{21}$), even though, using the Laplace transform approach, we determined 3 structurally identifiable parameter combinations. This is because the Laplace transform approach \emph{does} not account for the observability of the state $x_{2}$ whereas AutoRepar FISPO approach requires.

\subsection{Reparametrisation of CCM2 using COMBOS}
An alternative reparametrisation of the CCM2 model as found in \cite{meshkat2014COMBOS} is given by 
\begin{align*}
    \dot{\tilde{x}}_{1} &= -(-\Phi_{1} - \Phi_{3}) \tilde{x}_{1} - \Phi_{3} \tilde{x}_{1} + \tilde{x}_{2} \\
    \dot{\tilde{x}}_{2} &= \Phi_{3} \tilde{x}_{1} - (-\Phi_{2} - 1) \tilde{x}_{2} - \tilde{x}_{2},
\end{align*}
where
\begin{align*}
    \tilde{x}_{1} &= x_{1} \\
    \tilde{x}_{2} &= k_{12} x_{2}, \\
    \Phi_{1} &= -(k_{01} + k_{21}), \\
    \Phi_{2} &= -(k_{02} + k_{12}), \\
    \Phi_{3} &= k_{12} k_{21}.
\end{align*}
In this case, the reparametrisation is based on rewriting the original model equations in terms of 3 identifiable parameter combinations. The parameters of the resulting model are identifiable but the model is not FISPO since the state $\tilde{x}_{2}$ is not observable (determined using STRIKE GOLDD).

\begin{figure}[t!]
    \centering
    \includegraphics[width=0.48\textwidth]{
    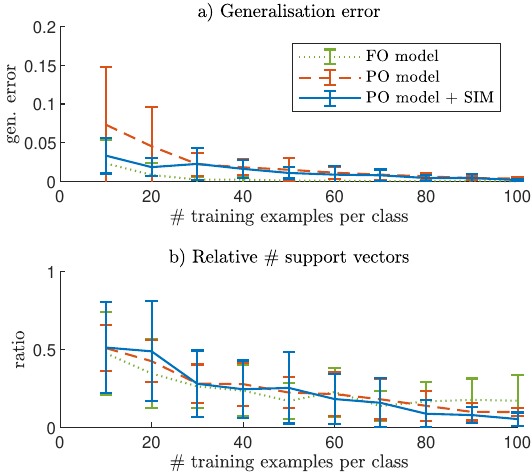}
    \caption{Experiment 1 with partially observed CCM2 model.}
    \label{fig:exp1_CCM2}
\end{figure}

\begin{figure}[t!]
    \centering
    \includegraphics[width=0.48\textwidth]{
    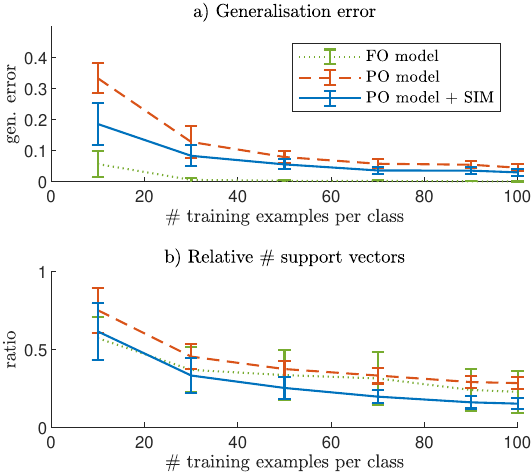}
    \caption{Experiment 1 with partially observed CCM4 model.}
    \label{fig:exp1_CCM4}
\end{figure}

\section{Identifiability Analysis for the CML}\label{app:SIA_CML}
Employing the Laplace transform approach on the CML, the following structurally identifiable parameter combinations have been determined:
\begingroup
\allowdisplaybreaks
\begin{align*}
      \tilde{\Phi}_{1} = & k_{04} k_{12} k_{23} + k_{12} k_{23} k_{34} + k_{04} k_{23} k_{42} + k_{04} k_{12} k_{43} +\\
               &k_{03} (k_{04} + k_{34}) (k_{12} + k_{42}) + k_{04} k_{42} k_{43} +  \\
               &k_{02} (k_{23} k_{34} + k_{03} (k_{04} + k_{34}) + k_{04} (k_{23} + k_{43})), \\
      \tilde{\Phi}_{2} = & k_{04} k_{12} + k_{04} k_{23} + k_{12} k_{23} + k_{12} k_{34} + k_{23} k_{34} +\\
               & k_{04} k_{42} + k_{23} k_{42} + k_{34} k_{42} +\\
               & k_{03} (k_{04} + k_{12} + k_{34} + k_{42}) + k_{04} k_{43} + k_{12} k_{43} +\\
               & k_{42} k_{43} + k_{02} (k_{03} + k_{04}+ k_{23} + k_{34} + k_{43}), +\\
      \tilde{\Phi}_{3} = & k_{02} + k_{03} + k_{04} + k_{12} + k_{23} + k_{34} + k_{42} + k_{43},\\
      \tilde{\Phi}_{4} = & k_{21} (k_{02} k_{23} k_{34} + k_{02} k_{03} (k_{04} + k_{34}) +\\
               & k_{03} (k_{04} + k_{34}) k_{42} + k_{02} k_{04} (k_{23} + \kb{k}_{43}) +\\
               & k_{04} k_{42} (k_{23} + k_{43})) + k_{01} (k_{04} k_{12} k_{23} +\\
               & k_{12} k_{23} k_{34} + k_{04} k_{23} k_{42} + k_{03} (k_{04} +\\
               & k_{34}) (k_{12} + k_{42}) + k_{04} k_{12} k_{43} + k_{04} k_{42} k_{43} +\\
               & k_{02} (k_{23} k_{34} + k_{03} (k_{04} + k_{34}) k_{04} (k_{23} + k_{43}))), \\
      \tilde{\Phi}_{5} = & k_{03} k_{04} k_{12} + k_{03} k_{04} k_{21} + k_{04} k_{12} k_{23} 
                 + k_{04} k_{21} k_{23} + \\
                 & k_{03} k_{12} k_{34} + k_{03} k_{21} k_{34} 
                 + k_{12} k_{23} k_{34} + k_{21} k_{23} k_{34} + \\
                 & k_{03} k_{04} k_{42} 
                 + k_{03} k_{21} k_{42} + k_{04} k_{21} k_{42} + k_{04} k_{23} k_{42} +\\
               & k_{21} k_{23} k_{42} + k_{03} k_{34} k_{42} + k_{21} k_{34} k_{42} 
               + k_{04} k_{12} k_{43} + \\
               & k_{04} k_{21} k_{43} + k_{04} k_{42} k_{43} 
               + k_{21} k_{42} k_{43} + k_{02} (k_{21} k_{23} + \\
               & k_{21} k_{34} + k_{23} k_{34} + 
               k_{03} (k_{04} + k_{21} + k_{34}) + k_{21} k_{43} + \\
               & k_{04} (k_{21} + k_{23} + k_{43})) 
               + k_{01} (k_{04} k_{12} + k_{04} k_{23} + \\
               & k_{12} k_{23} + k_{12} k_{34} + k_{23} k_{34} + 
               k_{04} k_{42} + k_{23} k_{42} + \\
               & k_{34} k_{42} + k_{03} (k_{04} + k_{12} + k_{34} + k_{42}) 
               + k_{04} k_{43} + \\
               & k_{12} k_{43} + k_{42} k_{43} 
               + k_{02} (k_{03} + k_{04} + k_{23} + k_{34} + k_{43})), \\
      \tilde{\Phi}_{6} = & k_{03} k_{04} + k_{03} k_{12} + k_{04} k_{12} + k_{03} k_{21} + k_{04} k_{21} +\\
               & k_{04} k_{23} + k_{12} k_{23} + k_{21} k_{23} + k_{03} k_{34} + k_{12} k_{34} +\\
               & k_{21} k_{34} + k_{23} k_{34} + k_{03} k_{42} + k_{04} k_{42} + k_{21} k_{42} +\\
               & k_{23} k_{42} + k_{34} k_{42} + k_{04} k_{43} + k_{12} k_{43} + k_{21} k_{43} +\\
               & k_{42} k_{43} + k_{02} (k_{03} + k_{04} + k_{21} + k_{23} + k_{34} + k_{43}) +\\
               & k_{01} (k_{02} + k_{03} + k_{04} + k_{12} + k_{23} + k_{34} + k_{42} + k_{43}), \\
      \tilde{\Phi}_{7} = & k_{01} + k_{02} + k_{03} + k_{04} + k_{12} + k_{21} \\
               & + k_{23} + k_{34} + k_{42} + k_{43}.
\end{align*}
\endgroup
Independent investigation using the Lie symmetry approach gave the simpler, yet equivalent, set of identifiable parameter combinations reported in Eq.~\eqref{eq:CML_SI} which were subsequently also used for experimentation.
A \textit{Wolfram Mathematica} script containing the corresponding analysis can be found on the authors' Github page. 

\section{Classification with a Support Vector Machine} \label{app:classification}
For the Support Vector Machine, the response to an input $x$ is modelled as
$$f(x) = \beta T(x) + b$$
where $\beta$ is a $p$-dimensional vector containing the weights to be determined, $b$ is a scalar bias term and $T(x)$ is implicitly determined via the choice of kernel function $K(\cdot, \cdot)$ and the relationship $K(x_{1},x_{2}) = T(x_{1}) \cdot T(x_{2})$. 
The Support Vector Machine is implemented with the MATLAB \texttt{fitcsvm} function. 
Default settings apply except for the \texttt{KernelFunction} setting, which is set to \texttt{gaussian} and the \texttt{Standardize} setting which is set to \texttt{true}.
Further, the default settings notably imply that: 1) the training employs Sequential Minimal Optimization, and 2) in the case of inseparable classes, slack variables $\xi_{j}$ are introduced and the objective becomes the minimization of 
$$\frac{1}{2} ||\beta||^2 + C \sum_{j=1}^{n}\xi$$
with respect to $\beta, b$ and $\xi_{j}$ subject to 
$$y_{j}f(x_{j}) \geq 1 - \xi_{j}, \quad \xi_{j} \geq 0. $$
The optimal settings for \texttt{BoxConstraint} and \texttt{KernelScale} are determined using a grid-search and 10-fold cross-validation to estimate out-of-sample performance.
Further information on the default settings of the MATLAB \texttt{fitcsvm} function can be found on the MATLAB \texttt{fitcsvm} documentation page.

\section{Additional experimental outcomes}\label{app:add_exp_outcomes}
\begin{figure}[t!]
    \centering
    \includegraphics[width=0.48\textwidth]{
    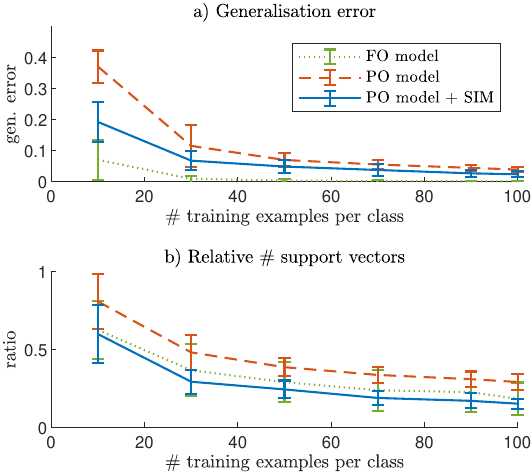}
    \caption{Experiment 1 with partially observed CML model.}
    \label{fig:exp1_CML}
\end{figure}

\begin{figure}[t!]
    \centering
    \includegraphics[width=0.48\textwidth]{
    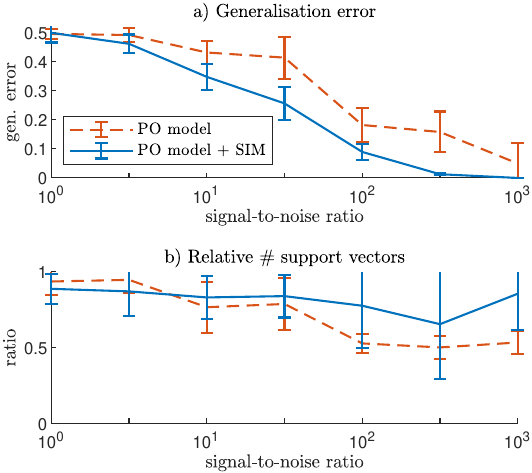}
    \caption{Experiment 2 with toy model.}
     \label{fig:exp2_toy}
\end{figure}

This appendix contains the experimental outcomes for the toy model, CCM2, CCM4, CML.
\autoref{fig:exp1_CCM2} through to \autoref{fig:exp1_CML} contain the results of experiment 1.
\autoref{fig:exp2_toy} through to \autoref{fig:exp2_CML} contain the results of experiment 2.
\autoref{fig:exp3_toy} through to \autoref{fig:exp3_CML} contain the results of experiment 3.
Note that the fully observed toy model is inherently unidentifiable and therefore experiment 1 has not been carried out for this model.

\begin{figure}[t!]
    \centering
    \includegraphics[width=0.48\textwidth]{
    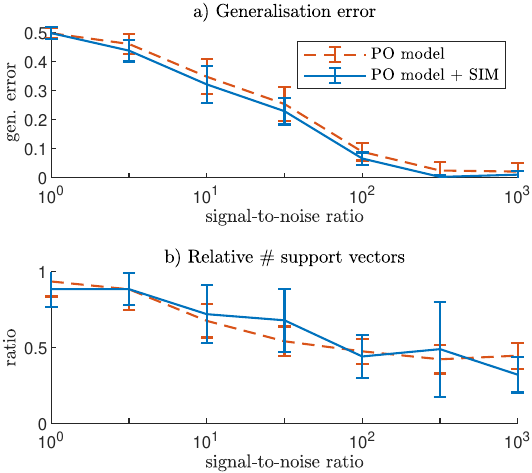}
    \caption{Experiment 2 with partially observed CCM2 model.}
    \label{fig:exp2_CCM2}
\end{figure}

\begin{figure}[t!]
    \centering
    \includegraphics[width=0.48\textwidth]{
    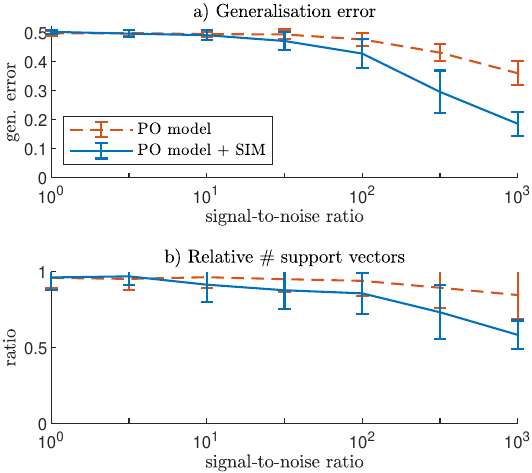}
    \caption{Experiment 2 with partially observed CCM4 model.}
    \label{fig:exp2_CCM4}
\end{figure}

\begin{figure}[t!]
    \centering
    \includegraphics[width=0.48\textwidth]{
    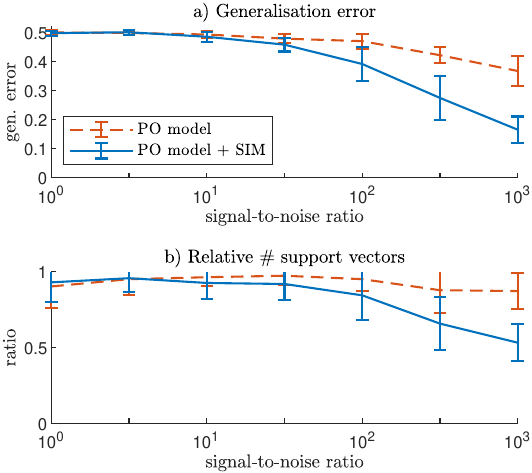}
    \caption{Experiment 2 with partially observed CML model.}
    \label{fig:exp2_CML}
\end{figure}


\begin{figure*}[t!]
    \centering
    \includegraphics[width=\textwidth]{
    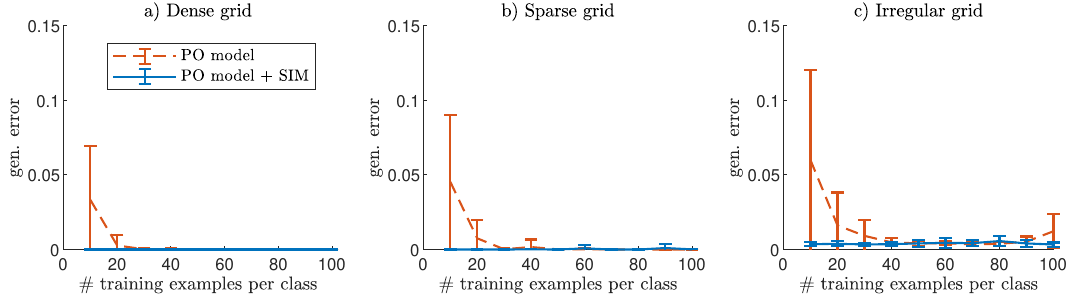
    }
    \caption{Experiment 3 with toy model.}
    \label{fig:exp3_toy}
\end{figure*}

\begin{figure*}[t!]
    \centering
    \includegraphics[width=\textwidth]{
    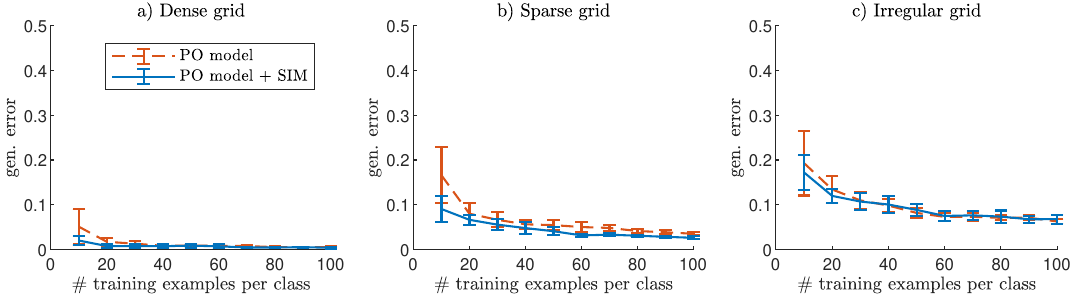
    }
    \caption{Experiment 3 with partially observed CCM2 model.}
    \label{fig:exp3_CCM2}
\end{figure*}

\begin{figure*}[t!]
    \centering
    \includegraphics[width=\textwidth]{
    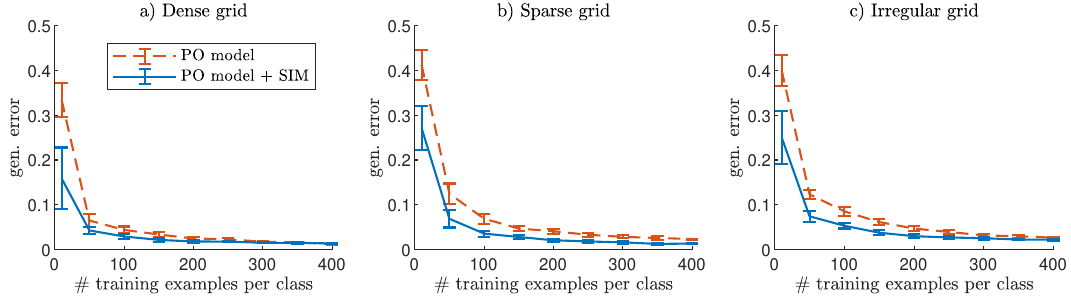
    }
    \caption{Experiment 3 with partially observed CCM4 model.}
    \label{fig:exp3_CCM4}
\end{figure*}

\begin{figure*}[t!]
    \centering
    \includegraphics[width=\textwidth]{
    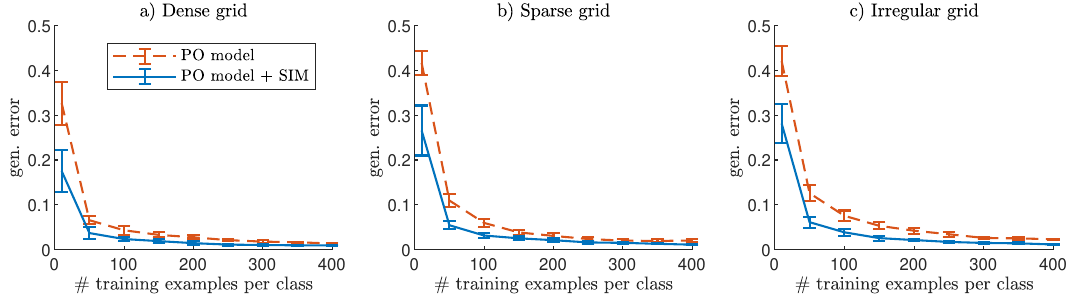
    }
    \caption{Experiment 3 with partially observed CML model.}
    \label{fig:exp3_CML}
\end{figure*}

\end{document}